\documentclass{article}
\usepackage{ijcai24}

\usepackage{times}
\usepackage{soul}
\usepackage{url}
\usepackage[hidelinks]{hyperref}
\usepackage[utf8]{inputenc}
\usepackage[small]{caption}
\usepackage{graphicx}
\usepackage{amsmath}
\usepackage{amsthm}
\usepackage{booktabs}
\usepackage{algorithm}
\usepackage{algorithmic}
\usepackage{amsfonts,amssymb} 
\usepackage{multirow}
\usepackage[switch]{lineno}
\usepackage[dvipsnames]{xcolor}
\usepackage{xcolor,colortbl}
\colorlet{tabfirst}{Green!25}
\definecolor{tabthird}{rgb}{1, 0.85, 0.7}
\definecolor{tabsecond}{rgb}{1, 0.96, 0.7}

\title{ HVOFusion: Incremental Mesh Reconstruction Using \\ Hybrid Voxel Octree}

\author{
Shaofan Liu$^1$
\and
Junbo Chen$^2$\and
Jianke Zhu$^1$\thanks{Corresponding authors}\\
\affiliations
$^1$Zhejiang University\\
$^2$Udeer.ai\\
\emails
\{liushaofan, jkzhu\}@zju.edu.cn,
junbo@udeer.ai
}

\begin{document}

\maketitle

\begin{abstract}
    Incremental scene reconstruction is essential to the navigation in robotics. Most of the conventional methods typically make use of either TSDF (truncated signed distance functions) volume or neural networks to implicitly represent the surface. Due to the voxel representation or involving with time-consuming sampling, they have difficulty in balancing speed, memory storage, and surface quality. In this paper, we propose a novel hybrid voxel-octree approach to effectively fuse octree with voxel structures so that we can take advantage of both implicit surface and explicit triangular mesh representation. Such sparse structure preserves triangular faces in the leaf nodes and produces partial meshes sequentially for incremental reconstruction. This storage scheme allows us to naturally optimize the mesh in explicit 3D space to achieve higher surface quality. We iteratively deform the mesh towards the target and recovers vertex colors by optimizing a shading model. Experimental results on several datasets show that our proposed approach is capable of quickly and accurately reconstructing a scene with realistic colors. Code is available at \small \url{https://github.com/Frankuzi/HVOFusion}
\end{abstract}

\section{Introduction}
Incremental scene reconstruction has achieved great progress in the recent years, which enables a wide range of applications, including motion planning~\cite{breitenmoser2012surface}, dense mapping~\cite{whelan2015real} and scene perception~\cite{hane20173d}. In contrast to offline scheme, the online reconstruction provides real-time feedback and adapt to dynamic scenes. Since the success of KinectFusion~\cite{izadi2011kinectfusion} with Truncated Signed Distance Function (TSDF) integration on a set of voxels~\cite{curless1996volumetric}, the underlying scene representation has seldom changed ever. It always relies on a large TSDF volume that usually costs a huge amount of memory, even for 3D indoor scenarios. Besides, the geometric accuracy of complex scenes~\cite{kim2010dynamic} is a bit limited due to the restrictions on voxel resolution.

\begin{figure}[t]
	\centering
	\includegraphics[width=0.48\textwidth]{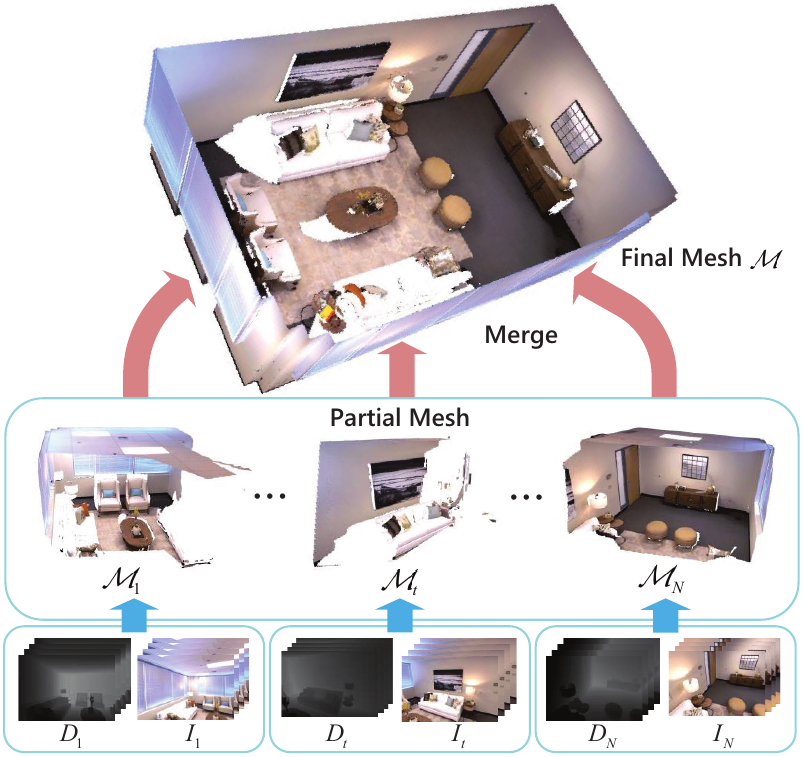}
	\caption{ \textbf{We reconstruct the scene in an incremental manner}. The partial mesh $\mathcal{M}_t$ output at time $t$ is constructed by the hybrid voxel-octree and deformed in the refinement branch. We merge all the partial meshes to get the final mesh $\mathcal{M}$.}
	\label{fig1}
\vspace{-2mm}
\end{figure}

On the other hand, recent research studies~\cite{ortiz2022isdf,zhu2022nice,yang2022vox} have demonstrated the promising results of implicit representation parameterized by neural networks. By formulating the scene surface as a continuous implicit function, its geometry can be extracted at arbitrary resolution. However, most of these methods are trained by minimizing the reconstruction loss of function values at a set of sampled 3D locations, which requires expensive sampling strategies to locate the surface and tends to render artifacts in synthesizing fine details. Recently, 3D Gaussian Splatting~\cite{kerbl3Dgaussians} attempts to represent the geometry by implicit ellipsoids in an explicit 3D space. The mesh and  attributes encoded in 3D Gaussians cannot be easily extracted for various downstream tasks.

A possible solution to solve these problems is to make use of explicit surface representation~\cite{shen2021deep,worchel2022multi,munkberg2022extracting,hanocka2020point2mesh,walker2023explicit}, which directly optimize mesh vertices coordinates and topological structures without intensive sampling. Therefore, it is able to trade-off between the efficiency and performance. Unfortunately, most of these methods make assumption of a given and fixed mesh topology, which are typically suitable for recovering individual objects. It is usually hard to incrementally reconstruct the scene geometry that does not have the specific boundaries.

To address the above challenges, we introduce HVOFusion, incremental mesh reconstruction using hybrid voxel-octree. The hybrid voxel-octree fuses octree and voxel structures, where the leaf nodes consist of a hierarchical voxel. This hybrid structure can achieve higher storage precision without increasing the octree depth. The key feature of this structure is that it leverages both implicit and explicit triangular mesh representations. Specifically, the leaf node predicts a coarse triangular mesh through an implicit function, where the resolution of the triangular mesh depends on the level of the voxel. The triangular mesh is explicitly stored in the leaf nodes of the octree so that the scene mesh can be incrementally reconstructed while the octree is being built. As shown in Fig.\ref{fig1}, this structure incrementally reconstructs the partial meshes of the scene with sequential input. This storage scheme allows us to naturally optimize the mesh in explicit 3D space to achieve higher surface quality. We refine the partial mesh extracted by the octree through image and point cloud supervision. Specifically, we iteratively deform the mesh towards the target, in which spherical harmonics are employed to model the vertex colors and ambient lighting. To the best of our knowledge, our method is the first to combine explicit surface optimization with online scene reconstruction. We conduct experiments on various datasets, whose promising results show the fast and accurate scene reconstruction with realistic colors.

The  main contributions are summarized as follows:
\begin{itemize}
\item An incremental mesh reconstruction using hybrid voxel-octree.
\item We propose the hybrid voxel-octree, which combines the advantages of octree and voxel structures, and integrates implicit and explicit triangular mesh representations.
\item We adopt images and point clouds as supervisions which allow us to effectively match the colors and lighting of the scene and directly enhance surface recovery.
\end{itemize}

\begin{figure*}[t]
	\centering
	\includegraphics[width=1\textwidth]{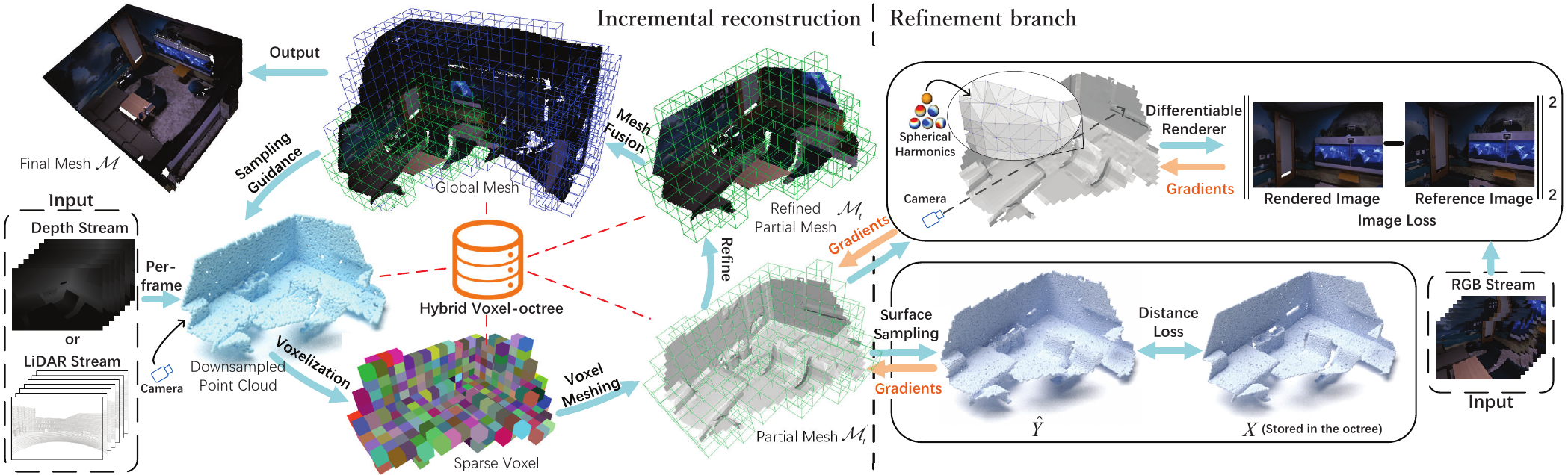}
	\caption{ \textbf{The architecture of our proposed incremental mesh reconstruction approach.} In the incremental reconstruction pipeline, each frame of the point cloud is downsampled and inserted into the hybrid voxel-octree, resulting in a sparse voxel structure composed of leaf nodes. The hybrid voxel-octree is then used to extract partial mesh ${\mathcal{M}_t'}$. ${\mathcal{M}_t'}$ is optimized in terms of its topology, colors, and vertex positions through the point-based and shading-based refinement branch. The refined partial mesh ${\mathcal{M}_t}$ is finally fused into the global mesh.}
	\label{fig2}
\vspace{-3mm}
\end{figure*}

\section{Related Work}
\subsection{TSDF-based Reconstruction}
A wide range of methods, including stereo matching and SLAM-based techniques, have been investigated for traditional scene reconstruction~\cite{engel2013semi,schonberger2016structure}. KinectFusion~\cite{izadi2011kinectfusion} is the seminal work on this task. Many following studies focused on volume-based methods, including Kintinuous~\cite{whelan2012kintinuous}, BundleFusion~\cite{dai2017bundlefusion}, InfiniTAM~\cite{prisacariu2017infinitam} and Voxgraph~\cite{reijgwart2019voxgraph}. Usually, the resolution of each voxel is often fixed in practice due to efficiency, which limits the capability of such representation. Other methods attempt to address this problem by representing the scene by a set of points or surfels~\cite{schops2019surfelmeshing}. The main drawback of point-based method is their discrete nature, which does not have the important topology information. In this work, we employ the hybrid voxel-octree structure to store triangular faces so that the explicit meshes can be deformed to the target scene. Thus, our method are able to achieve higher reconstruction accuracy while consuming less memory.

\subsection{Neural Scene Reconstruction}
~\cite{mildenhall2021nerf} represent the scene as a neural radiance field, which reconstruct indoor scenes through implicit neural representations. Later, iMAP~\cite{sucar2021imap} performs both tracking and mapping using neural implicit representations. To improve the scalability, NICE-SLAM~\cite{zhu2022nice} suggest a hierarchical multi-feature grids. Several follow-up works have improved upon these approaches from various perspectives. ESLAM~\cite{johari2023eslam}, iDF-SLAM~\cite{ming2022idf}, and iSDF~\cite{ortiz2022isdf} leverage surface rendering-based methods to improve the accuracy of reconstruction. Recently, 3D Gaussian splatting~\cite{kerbl3Dgaussians} optimizes and renders scenes very efficiently with high quality. SuGaR~\cite{guedon2023sugar} extracts an accurate mesh from the Gaussians. SplaTAM~\cite{keetha2023splatam} simultaneously estimates the camera poses and fit the underlying Gaussian splatting.

\subsection{Explicit Surface Representation}
Explicit surface reconstruction methods aim to estimate 3D triangular mesh from images. Meshes in a scene are firstly rasterized to create a screen-space map containing geometric information. The raster images are then processed by a shader to predict pixel-wise RGB values. These methods~\cite{wang2018pixel2mesh,liu2019soft} have achieved impressive results in reconstructing and synthesizing simple shapes. DefTet~\cite{gao2020learning} represents a mesh with a deformable tetrahedral grid where the grid vertex coordinates and the occupancy values are learned from the neural network. DMTet~\cite{shen2021deep} optimizes the surface mesh directly through a differentiable marching tetrahedral layer.~\cite{thies2019deferred} fully parameterize a deferred shader by a neural network, and hence synthesize novel views with arbitrary illumination and materials. In spite of the promising results, these methods are limited to smaller objects, which are hard to scale into larger scenes. To this end, our proposed approach incrementally reconstructs the large scene and refines each partial reconstruction individually.

\section{Method}
In contrast to the conventional offline methods, we aim to reconstruct the large scale scene in an online manner, as shown in Fig.\ref{fig1}. Give a stream of input image $I_t$ with the depth $D_t$ or the point cloud $P_t$, we reconstruct a partial mesh $\mathcal{M}_t = (\mathcal{V}, \mathcal{E}, \mathcal{F})$, which consists of vertex positions $\mathcal{V}$, a set of edges $\mathcal{E}$, and a set of faces $\mathcal{F}$. The final mesh $\mathcal{M}$ is composed of $N$ partial meshes. The partial mesh is extracted and stored by hybrid voxel-octree (Sec.~\ref{sec:octree}). We optimize partial mesh by point-based refinement (Sec.~\ref{sec:point}) and shading-based refinement (Sec.~\ref{sec:shading}) to obtain refined partial mesh. Fig.\ref{fig2} illustrates the overall framework of our presented method.

\subsection{Hybrid Voxel-octree}
\label{sec:octree}
\subsubsection{Hybrid Voxel-octree Surface Representation}
Generally, it is easy to implement the voxel-based surface representation, which enables the parallel data access. However, it often consumes large amount of memory. Instead, the octree-based surface representation offers high spatial efficiency while their complexity increases along with the depth. 

To tackle these issues, we propose a novel hybrid voxel-octree structure that is able to take advantages of these two kinds of representations. Instead of using the single list like the typical octree, the leaf nodes of our voxel-octree are composed of the hierarchical voxel, which can achieve higher storage precision without increasing the octree depth. As shown in Fig.\ref{fig3}, we insert the points into the voxel at the leaf node of voxel-octree. Then, we compute the normal vector for each point to determine the average curvature of the voxel. When the average curvature is above the threshold $\mathcal{T}_{cur}$, we subdivide it into ${L^3}$ multi-layered voxels. $L$ represents the level of the voxel. Such scheme enables to generate the more detailed triangular faces without increasing the octree depth or reinserting points. During the construction of the octree, we extract the triangulated mesh on the leaf nodes to obtain the partial surface reconstruction. It is worthy of noting that we only store the vertices and topology of the triangular faces, which greatly reduces the memory consumption.

\subsubsection{Voxelization}
\begin{figure}
	\centering
	\includegraphics[width=0.40\textwidth]{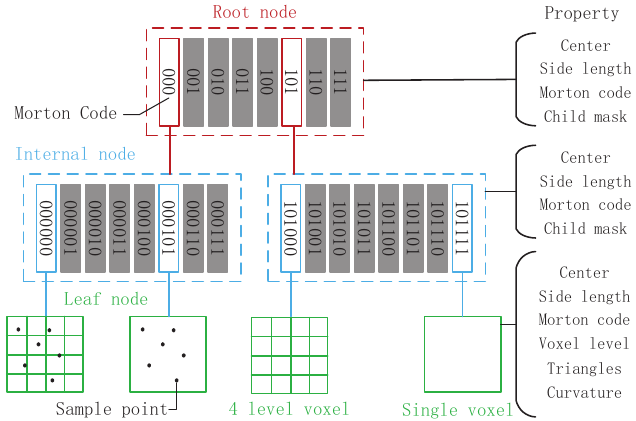}
	\caption{ \textbf{Example of the structure and node properties of the hybrid voxel-octree.} Each node is indicated by a binary Morton code representing its position. }
	\label{fig3}
\vspace{-2mm}
\end{figure}

The sparse point cloud is obtained from either LiDAR scanner or the back-projection of depth map. We insert each point into the leaf nodes of the octree, which is further stored in the hierarchical voxel. To reduce the computational complexity, the point cloud is downsampled at a certain ratio in the insertion process. The minimum $\mathcal{L}_1$ distance from the current point to the other points within the voxel is employed to determine whether the point could be inserted into the voxel. In our implemntation, we choose the points whose $\mathcal{L}_1$ distances fall below the threshold $\mathcal{T}_{min}$. As shown in Fig.~\ref{fig3}, we encode voxel coordinates as Morton codes for fast voxel indexing, which is able to locate voxel of leaf node and its $26$ adjacent nodes in the time complexity of $O(1)$.

\subsubsection{Voxel Meshing}
\begin{figure}
	\centering
	\includegraphics[width=0.39\textwidth]{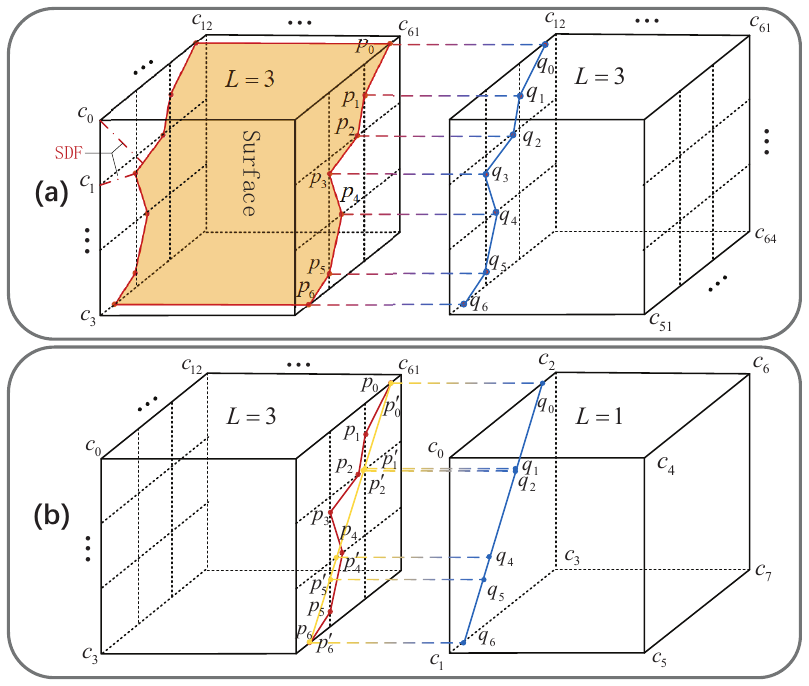}
	\caption{ \textbf{Example of voxel meshing.} $c_n$ is the voxel's corner related to voxel level. $p_n$ and $q_n$ are vertices of triangular face within the voxel. a). The positions of $p_n$ and $q_n$ are consistent when adjacent voxels are at the same level b). To ensure consistency, $p_n$ is modified to $p'_n$ when adjacent voxels are at different levels. }
	\label{fig7}
\vspace{-2mm}
\end{figure}

For each leaf node in the octree, the goal of voxel meshing is to find a zero level-set $\mathcal{S}$ within the voxel, which is equivalent to the implicit surface to be recovered.  Unlike TSDF-based methods that predict only one TSDF value for each voxel, as shown in Fig.\ref{fig7}, we treat each voxel as a cube and compute the signed distance function (SDF) by measuring the distance from its cube corners to the zero level-set
\begin{equation}
    SDF({p_i}) = \mathbf{n}_q^T(q - {p_i}),i \in [0,{(L + 1)^3})
\end{equation}
where $p_i$ represents the corner of the cube. $q$ denotes the nearest neighbor of $p_i$, and $\mathbf{n}_q$ represents the normal vector of the point $q$. The number of corner point $p_i$ is determined by the voxel’s level $L$.

The TSDF-based methods, such as voxel hashing~\cite{niessner2013real}, typically employ the Marching Cubes algorithm ~\cite{lorensen1998marching} to extract triangulated mesh from a large TSDF volume stored in hashing. However, these methods usually adopt voxels of the same size to ensure that the triangular faces can be correctly extracted from the adjacent 8 voxels. To improve the surface accuracy, they use smaller voxel size to increase the number of triangular faces, which causes inefficient computations for regions with simple geometry. In contrast, we suggest a sparse voxel Marching Cubes method, which supports extracting triangular faces from voxels with different levels. This method is able to achieve more accurate surface reconstruction without changing the octree structure.

Specifically, when extracting triangular faces within the voxel, the vertices of triangular faces are located on the boundaries of the voxels. As shown in Fig.\ref{fig7}, the positions of the vertices $p_0$ to $p_6$ and $q_0$ to $q_6$ are obviously consistent when the adjacent voxels are at the same level. This is because the adjacent voxels have the same corner division at the connection. When adjacent voxels are at different levels, we use the vertices $p_0$ to $p_6$ of the high-level voxel to linearly approximate the vertices $q_0$ to $q_6$ of the low-level voxel, resulting in vertices $p'_0$ to $p'_6$. This ensures the consistency of the vertex positions between any voxels.

\subsection{Point-based Refinement}
\label{sec:point}
The TSDF-based reconstruction methods~\cite{izadi2011kinectfusion,niessner2013real} may have difficulty accurately representing complex and intricate regions due to the restrictions on voxel resolution. Instead, by directly preserve the triangular faces, we can refine the mesh in explicit 3D space to overcome this limitation. We iteratively deform the partial mesh generated by the hybrid voxel-octree towards the input point cloud $X$. The point cloud $X$ is stored in the octree, which effectively preserves the detailed scene geometry. Specifically, we optimize the differential vertex position $\Delta v_{t}$ of the mesh rather than its absolute coordinates. The deformed vertex position can be represented as $v_{t}=v_{t-1}+\Delta v_{t}$. In contrast the conventional methods~\cite{sharf2006competing,hanocka2020point2mesh} learning the priors through neural networks, we directly treat the partial mesh as an explicit geometric prior, which enables the rapid convergence of vertices towards their optimal positions and without any neural networks. The vertices of refined partial mesh ${M_t}$ are driven by the distance of the partial mesh ${M'_t}$ to the input point cloud $X$. We sample the deformed mesh ${M'_t}$ to obtain the sampled points $\hat{Y}$. Then, we minimize the Chamfer distance between point cloud $X$ and $\hat{Y}$ as follows
\begin{equation}
    {\cal L}_{p o i n s} = \sum_{x\in X}\operatorname*{min}_{{\hat{y}}\in{\hat{Y}}}\|x-{\hat{y}}\|_{2}+\sum_{{\hat{y}}\in{\hat{Y}}}\operatorname*{min}_{x\in X}\|x-{\hat{y}}\|_{2}
\end{equation}
In this work, we adopt the same differential sampling mechanism like~\cite{hanocka2020point2mesh}. The distance between the sampled points and the corresponding point cloud contributes to a gradient update, which guides the update of the three vertices coordinates of each triangle.

\subsection{Shading-based Refinement}
\label{sec:shading}
\subsubsection{Lighting Estimation}
As in~\cite{wu2011high}, we learn the vertex colors of the triangular mesh and refine the surface through optimizing the lighting model. Our method is able to recover more realistic colors compared to implicit methods that calculate vertex colors through integration of cumulative color. In general, the irradiance equation for each vertex can be defined as below
\begin{equation}
    B(i,\mathbf{n}) = \rho (i)S(\mathbf{n}) + \beta (i)
    \label{eq5}
\end{equation}
where $B(i,{\mathbf{n}})$ represents the image lighting at each vertex. $S(\mathbf{n})$ denotes the shading, and $\rho{(i)}$ is the vertex albedo to adjust the local shadow intensity. $\beta{(i)}$ accounts for local lighting variations. For simplicity, $\beta{(i)}$ is set to zero during the training. We describe the local lighting and shadows as low order spherical harmonic. Therefore, the shading can be formulated as a linear polynomial of the normal vector and spherical harmonic coefficients
\begin{equation}
    S(\mathbf{n}) = \sum\limits_{k = 1}^4 {{l_k}} \mathbf{n}
    \label{eq6}
\end{equation}
where $l$ represents a vector of the four first order spherical harmonics coefficients. $\mathbf{n}$ denotes vertex normal.

\subsubsection{Albedo Recovery}
The shading estimation in Eq.\ref{eq6} is a rough assessment of the lighting, which only holds for diffuse surfaces with uniform albedo. It has difficulties in handling the specularities, shadows, and nearby light sources that may exist in the real-world scenarios. Therefore, we simultaneously optimize both the albedo and the spherical harmonic in Eq.\ref{eq5}. Similar to~\cite{worchel2022multi}, we adopt a differentiable deferred shading pipeline to render into images, which is inspired by real-time graphics. Given the camera parameter, the mesh is rasterized to yield the per pixel triangle index and barycentric coordinates, which are further employed to interpolate both the vertex albedo and vertex spherical harmonics. Our goal is to minimize the differences between the rendered image and the input frame as follows
\begin{equation}
    {\cal L}_{shading} = \left\| {B(i,\mathbf{n}) - I} \right\|_2^2
\end{equation}
Since the gradient of ${\cal L}_{shading}$ can be backpropagated to the mesh vertices, we are able to recover the details of the surface geometry and the corresponding vertex colors.

\begin{figure*}[t]
	\centering
	\includegraphics[width=1\textwidth]{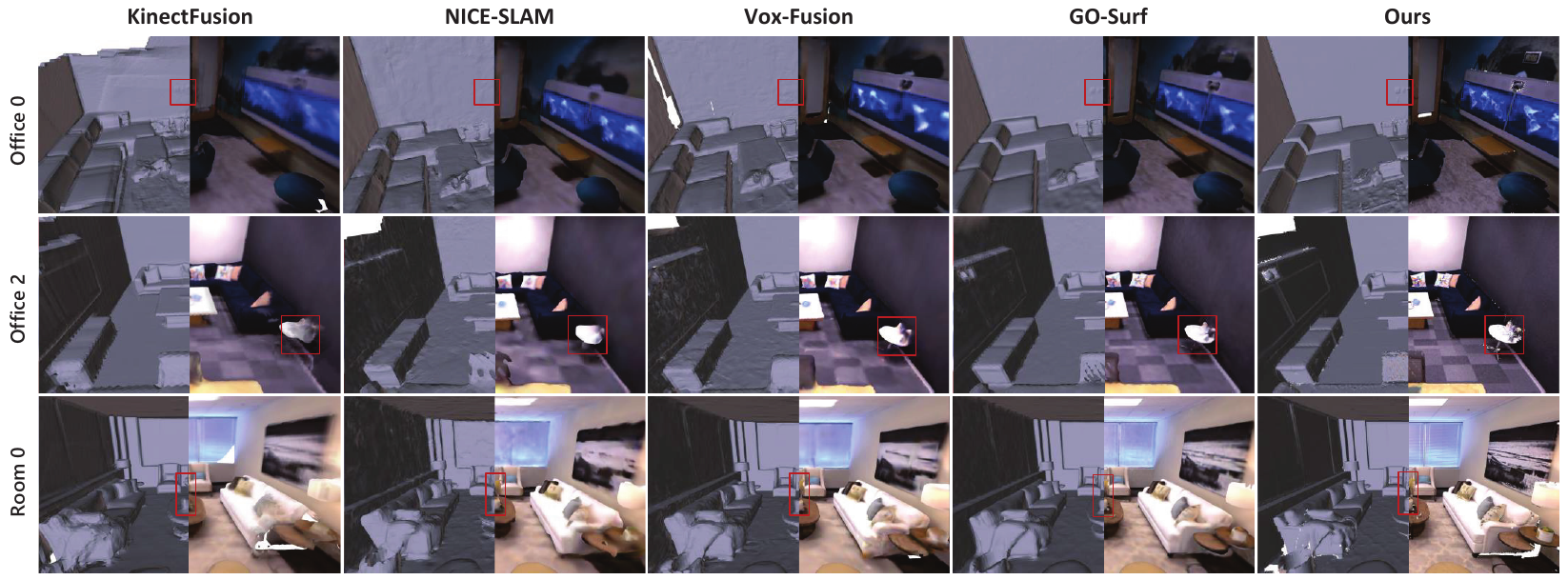}
	\caption{ \textbf{Reconstruction Result for Replica Dataset.} The right half shows the result after vertex coloring. Our method recovers more accurate and smoother geometric shapes than implicit methods, particularly in flat and detailed regions. }
	\label{fig4}
\vspace{-2mm}
\end{figure*}

\subsection{Objective Function}
Directly deforming the mesh may lead to the undesirable meshes with degenerated triangles and self-intersections. To tackle this issue, the geometric regularization~\cite{luan2021unified} is typically employed. 

Let $V$ denote an $n\times 3$ matrix with vertex positions as rows, and $L\in\mathbb{R}^{n\times n}$ is the Laplacian matrix. Thus, the Laplacian regularization term is defined as follows
\begin{equation}
    {\cal L}_{lap} = {\lambda _{lap}}{\left\| {LV} \right\|^2}
\end{equation}

The normal consistency term  enforces the surface smoothness by calculating the cosine similarity between the neighboring face normals as follows
\begin{equation}
    {\cal L}_{normal} = {\lambda _{{normal}}}{\Sigma _{i,j}}{[1 - ({\mathbf{n}_i} \cdot {\mathbf{n}_j})]^2}
\end{equation}
where the pair$(i,j)$ represents the $i$-th and the $j$-th triangles sharing a common edge. 

The edge length loss penalizes the long edge in the mesh as follows
\begin{equation}
    {\cal L}_{edge} = {\lambda _{{edge}}}{({\Sigma _i}e_i^2)^{1/2}}
\end{equation}
where $e_i$ denotes the length of the $i$-th face edge.

Finally, the overall objective function can be formulated as follows
\begin{equation}
    {\cal L}={\cal L}_{p o i n s}+{\cal L}_{s h a d i n g}+{\cal L}_{l a p}+{\cal L}_{n o r m a l}+{\cal L}_{e d g e}
\end{equation}

\section{Experiments}
In this section, we conduct a series of experiments to evaluate the reconstruction and rendering quality of our proposed approach. Please refer to our supplementary materials for more ablation experiments and visualization results.
\subsection{Experimental Setup}
\subsubsection{Datasets and Baselines}
We conduct experiments on three different kinds of datasets, including Replica~\cite{replica19arxiv}, ScanNet++~\cite{yeshwanthliu2023scannetpp}, and Newer College Dataset~\cite{ramezani2020newer}. They represent, an indoor synthetic dataset, a real indoor scene dataset, and an outdoor 3D LiDAR dataset, respectively.

To demonstrate the efficacy of our proposed approach, the following methods are employed as baseline: 1) KinectFusion~\cite{izadi2011kinectfusion} incrementally fuses the consecutive frames of depth data into a 3D volumetric representation of an implicit surface; 2) NICE-SLAM~\cite{zhu2022nice} incorporates multi-level local information via a hierarchical scene representation; 3) Vox-Fusion~\cite{yang2022vox} combines both neural and volumetric methods; 4) SplaTAM~\cite{keetha2023splatam} represents a scene by 3D Gaussian Splatting using a single unposed monocular RGB-D camera; 5) GO-Surf~\cite{wang2022go-surf} combines a multi-resolution feature grid with a hybrid volume rendering scheme; 6) Puma~\cite{vizzo2021icra} employs the Poisson surface reconstruction to obtain triangular mesh from the LiDAR scan. To facilitate fair evaluation on surface reconstruction, we directly make use of ground truth poses for all baselines in our experiments.

\begin{table}[t]
\centering
\scriptsize
\setlength{\tabcolsep}{1.2pt}

\begin{tabular}{lcccccccccc}
\toprule
\textbf{Methods} & \textbf{Metrics} & \texttt{R0} & \texttt{R1} & \texttt{R2} & \texttt{O0} & \texttt{O1} & \texttt{O2} & \texttt{O3} & \texttt{O4} & \textbf{Avg.} \\
\midrule

\multirow{3}{*}{KinectFusion}
& Acc {[}cm{]}↓ & 12.85  & 22.44  & 21.46  & 11.67    & 10.44    & 14.77    & 13.87    & 15.23    & 15.34 \\
& NC↑           & 0.83   & 0.76   & 0.78   & 0.77     & 0.77     & 0.83     & 0.82     & 0.78     & 0.79  \\
& F-score↑      & 0.28   & 0.25   & 0.20   & 0.37     & 0.42     & 0.37     & 0.33     & 0.40     & 0.33  \\

\hline

\multirow{3}{*}{NICE-SLAM}
& Acc {[}cm{]}↓ & \cellcolor{tabthird}2.73   & 2.58   & \cellcolor{tabsecond}2.65   & \cellcolor{tabthird}2.26     & \cellcolor{tabthird}2.50     & 3.82     & \cellcolor{tabthird}3.50     & \cellcolor{tabsecond}2.77     & \cellcolor{tabthird}2.85  \\
& NC↑           & 0.92   & 0.79   & 0.91   & 0.90     & 0.88     & 0.88     & 0.86     & \cellcolor{tabthird}0.91     & 0.88  \\
& F-score↑      & \cellcolor{tabsecond}0.90   & \cellcolor{tabthird}0.52   & \cellcolor{tabsecond}0.90   & \cellcolor{tabsecond}0.88     & 0.84     & 0.83     & 0.69     & \cellcolor{tabthird}0.90     & 0.81  \\

\hline

\multirow{3}{*}{Vox-Fusion}
& Acc {[}cm{]}↓ & \cellcolor{tabsecond}2.41   & \cellcolor{tabsecond}1.62   & \cellcolor{tabthird}3.11   & \cellcolor{tabsecond}1.74     & \cellcolor{tabsecond}1.69     & \cellcolor{tabsecond}2.23     & \cellcolor{tabsecond}2.84     & \cellcolor{tabthird}3.31     & \cellcolor{tabsecond}2.37  \\
& NC↑           & \cellcolor{tabsecond}0.94   & \cellcolor{tabthird}0.91   & \cellcolor{tabthird}0.91   & \cellcolor{tabthird}0.91     & \cellcolor{tabthird}0.89     & \cellcolor{tabthird}0.92     & \cellcolor{tabsecond}0.91     & 0.90     & \cellcolor{tabthird}0.91  \\
& F-score↑      & \cellcolor{tabfirst}0.93   & \cellcolor{tabsecond}0.81   & 0.83   & \cellcolor{tabsecond}0.88     & \cellcolor{tabsecond}0.87     & \cellcolor{tabsecond}0.88     & \cellcolor{tabsecond}0.87     & \cellcolor{tabsecond}0.91     & \cellcolor{tabsecond}0.87  \\

\hline

\multirow{3}{*}{GO-Surf}
& Acc {[}cm{]}↓ & 5.16   & \cellcolor{tabthird}2.49   & 7.13   & 4.40     & 3.60     & \cellcolor{tabthird}3.48     & 5.32     & 5.52     & 4.64  \\
& NC↑           & \cellcolor{tabthird}0.93   & \cellcolor{tabsecond}0.91   & \cellcolor{tabsecond}0.93   & \cellcolor{tabsecond}0.94     & \cellcolor{tabsecond}0.93     & \cellcolor{tabsecond}0.93     & \cellcolor{tabthird}0.89     & \cellcolor{tabsecond}0.92     & \cellcolor{tabsecond}0.92  \\
& F-score↑      & \cellcolor{tabthird}0.89   & \cellcolor{tabsecond}0.81   & \cellcolor{tabthird}0.85   & \cellcolor{tabthird}0.85     & \cellcolor{tabthird}0.84     & \cellcolor{tabthird}0.86     & \cellcolor{tabthird}0.85     & 0.87     & \cellcolor{tabthird}0.85  \\

\hline

\multirow{3}{*}{Ours}
& Acc {[}cm{]}↓ & \cellcolor{tabfirst}\textbf{0.57}   & \cellcolor{tabfirst}\textbf{0.56}   & \cellcolor{tabfirst}\textbf{0.57}   & \cellcolor{tabfirst}\textbf{0.53}     & \cellcolor{tabfirst}\textbf{0.55}     & \cellcolor{tabfirst}\textbf{0.58}     & \cellcolor{tabfirst}\textbf{0.59}     & \cellcolor{tabfirst}\textbf{0.57}     & \cellcolor{tabfirst}\textbf{0.56}  \\
& NC↑           & \cellcolor{tabfirst}\textbf{0.97}   & \cellcolor{tabfirst}\textbf{0.96}   & \cellcolor{tabfirst}\textbf{0.97}   & \cellcolor{tabfirst}\textbf{0.94}     & \cellcolor{tabfirst}\textbf{0.94}     & \cellcolor{tabfirst}\textbf{0.95}     & \cellcolor{tabfirst}\textbf{0.94}     & \cellcolor{tabfirst}\textbf{0.93 }    & \cellcolor{tabfirst}\textbf{0.94}  \\
& F-score↑      & \cellcolor{tabfirst}\textbf{0.93}   & \cellcolor{tabfirst}\textbf{0.94}   & \cellcolor{tabfirst}\textbf{0.93}   & \cellcolor{tabfirst}\textbf{0.89}     & \cellcolor{tabfirst}\textbf{0.88}     & \cellcolor{tabfirst}\textbf{0.91}     & \cellcolor{tabfirst}\textbf{0.93}     & \cellcolor{tabfirst}\textbf{0.93}     & \cellcolor{tabfirst}\textbf{0.92}  \\

\bottomrule
\end{tabular}
\caption{ \textbf{Quantitative evaluation of the reconstruction quality on the Replica dataset of 8 synthetic scenes.}  }
\label{tab1}
\end{table}

\subsubsection{Implementation Details}
Our presented hybrid voxel-octree method is implemented by C++. For indoor scenes, the edge length of the voxel-octree leaf nodes is set to $0.05$ $m$. The threshold $\mathcal{T}_{min}$ is $0.02$ $m$, and the threshold $\mathcal{T}_{cur}$ is set to $0.01$. The refinement process is implemented by PyTorch with the ADAM optimizer. The learning rates for vertex positions, vertex colors, and spherical harmonics are set to $0.0001$, $0.01$, and $0.001$, respectively. The weights for the loss terms are set to $50$ for $\lambda _{lap}$, $1$ for $\lambda _{normal}$ and $1$ for $\lambda _{edge}$. We run 300 iterations to optimize each partial mesh. The entire pipeline ran in an end-to-end fashion without requiring any network or pre-training. All experiments are conducted on a desktop PC with an NVIDIA RTX3090 GPU.

\begin{figure}[t]
	\centering
	\includegraphics[width=0.48\textwidth]{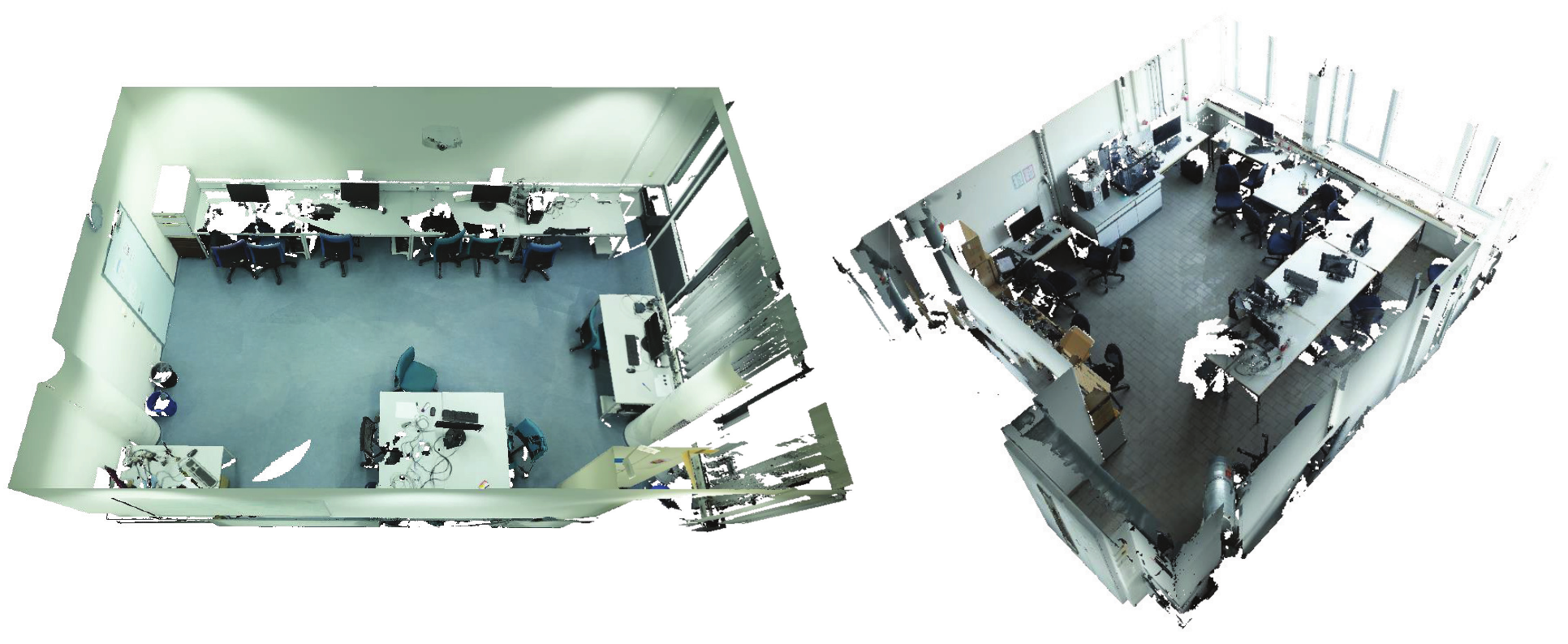}
	\caption{ \textbf{Other qualitative textured results on ScanNet++ reconstructed online with our method.} }
	\label{fig10}
\end{figure}

\begin{figure}[t]
	\centering
	\includegraphics[width=0.48\textwidth]{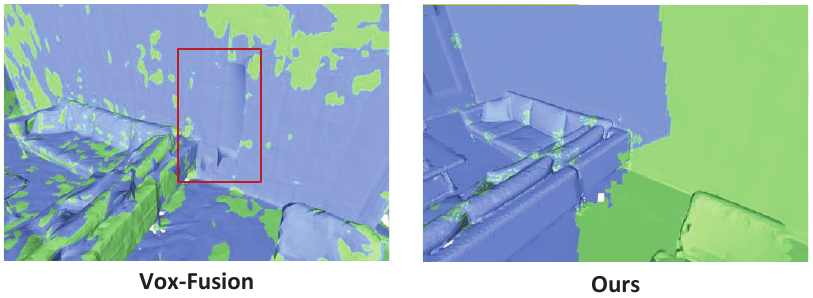}
	\caption{ \textbf{Visualization results of mesh edges during incremental reconstruction.} The blue and green regions represent two distinct partial meshes, and the red box indicates the artifacts produced by Vox-Fusion during mesh fusion. Our meshes have minimal overlapping triangles and accurate edge alignment. }
	\label{fig5}
\end{figure}

\subsection{Evaluation on Surface Reconstruction}
We investigate the reconstruction results on the Replica dataset. We measure accuracy, normal consistency and F-score to evaluate the reconstruction quality. The metrics are computed between point clouds sampled at a density of $1$ point per $cm^2$. F-score is computed using a threshold of $5$ $cm$. Fig.~\ref{fig4} shows that our method recovers more detailed geometry, especially retains the smoothness of the wall and floor surfaces. This is because we directly optimize the triangular face coordinates to enable a more faithful representation of the geometric intricacies in the scene. Our shading model also enhances the accuracy and richness of the surface texture and color. Tab.\ref{tab1} shows that our presented method surpasses all the other baselines in terms of accuracy, normal consistency, and F-score metrics.

We further compare our incremental reconstruction results with Vox-Fusion. We export and fuse partial meshes every 200 frames on the Replica dataset to simulate the runtime mesh output. As shown in Fig.\ref{fig5}, our meshes have minimal overlapping triangles and accurate edge alignment. In contrast, Vox-Fusion suffers from the overlapped triangles and surface intersections.

\begin{figure}[t]
	\centering
	\includegraphics[width=0.48\textwidth]{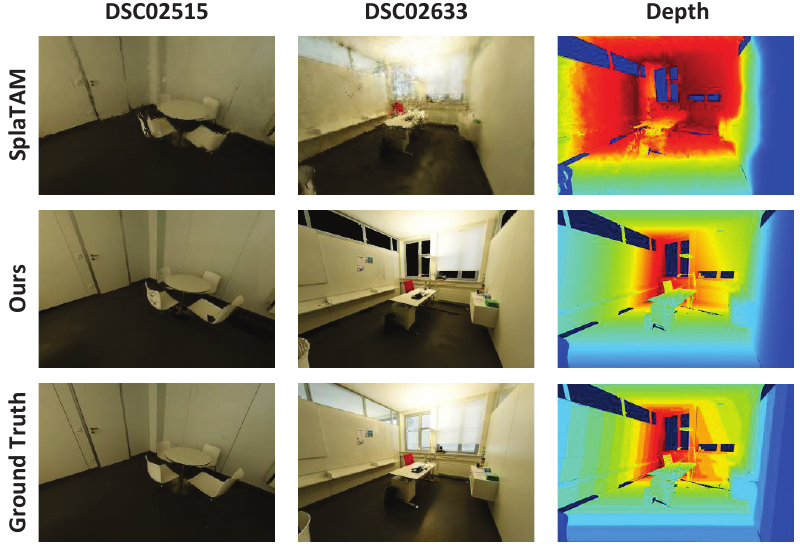}
	\caption{ \textbf{Evaluation of Rendering Quality on the ScanNet++ Dataset.} Qualitative results demonstrate that our method produces higher-quality rendering results compared to the baselines. }
	\label{fig6}
\end{figure}

\begin{table}[t]
\centering
\scriptsize
\setlength{\tabcolsep}{4.5pt}

\begin{tabular}{lccccccc}
\toprule
\multirow{2}{*}{\textbf{Methods}} & \multirow{2}{*}{\textbf{Metric}} & \multicolumn{3}{c}{\textbf{Novel View}} & \multicolumn{3}{c}{\textbf{Training View}} \\
&                         & S1       & S2       & \textbf{Avg.}     & S1        & S2        & \textbf{Avg.}      \\
\midrule
\multirow{4}{*}{SplaTAM} & PSNR↑                   & \cellcolor{tabsecond}23.99    & \cellcolor{tabsecond}24.48    & \cellcolor{tabsecond}24.41    & \cellcolor{tabsecond}27.82     & \cellcolor{tabfirst}\textbf{28.14}     & \cellcolor{tabsecond}27.98     \\
                         & SSIM↑                   & \cellcolor{tabsecond}0.88     & \cellcolor{tabfirst}0.87     & \cellcolor{tabfirst}0.88     & \cellcolor{tabfirst}\textbf{0.94}      & \cellcolor{tabfirst}\textbf{0.94 }     & \cellcolor{tabfirst}\textbf{0.94}      \\
                         & LPIPS↓                  & \cellcolor{tabsecond}0.21     & \cellcolor{tabsecond}0.26     & \cellcolor{tabsecond}0.24     & \cellcolor{tabfirst}\textbf{0.12}      & \cellcolor{tabfirst}0.13      & \cellcolor{tabfirst}\textbf{0.12}      \\
                         & Depth L1 {[}cm{]}↓      & \cellcolor{tabsecond}1.91     & \cellcolor{tabsecond}2.23     & \cellcolor{tabsecond}2.07     & \cellcolor{tabfirst}\textbf{0.93}      & \cellcolor{tabsecond}1.64      & \cellcolor{tabfirst}\textbf{1.28}      \\
\hline
\multirow{4}{*}{Ours}    & PSNR↑                   & \cellcolor{tabfirst}\textbf{26.64}    & \cellcolor{tabfirst}\textbf{25.43}    & \cellcolor{tabfirst}\textbf{26.04}    & \cellcolor{tabfirst}\textbf{28.06}     & \cellcolor{tabsecond}28.05     & \cellcolor{tabfirst}\textbf{28.05}     \\
                         & SSIM↑                   & \cellcolor{tabfirst}\textbf{0.89}     & \cellcolor{tabfirst}\textbf{0.87}     & \cellcolor{tabfirst}\textbf{0.88}     & \cellcolor{tabsecond}0.93      & \cellcolor{tabsecond}0.93      & \cellcolor{tabsecond}0.93      \\
                         & LPIPS↓                  & \cellcolor{tabfirst}\textbf{0.18}     & \cellcolor{tabfirst}\textbf{0.24}     & \cellcolor{tabfirst}\textbf{0.21}     & \cellcolor{tabsecond}0.13      & \cellcolor{tabfirst}\textbf{0.13}      & \cellcolor{tabsecond}0.13      \\
                         & Depth L1 {[}cm{]}↓      & \cellcolor{tabfirst}\textbf{1.84}     & \cellcolor{tabfirst}\textbf{1.99}     & \cellcolor{tabfirst}\textbf{1.92}     & \cellcolor{tabsecond}1.53      & \cellcolor{tabfirst}\textbf{1.48}      & \cellcolor{tabsecond}1.51      \\
\bottomrule
\end{tabular}
\caption{ \textbf{Novel \& Train View Rendering Performance on ScanNet++.} Our method is on par in terms of performance with SplaTAM but exhibits better results in novel view rendering. }
\label{tab6}
\end{table}

\subsection{Evaluation on Rendering Quality}
We evaluate the rendering quality on the input views from the ScanNet++ dataset in Tab.\ref{tab6}. We employ thee same evaluation metrics as in~\cite{keetha2023splatam}, including PSNR, SSIM and LPIPS. We evaluate the presented method on two scenes (8b5caf3398 (S1) and b20a261fdf (S2)). It can be seen that our approach performs comparably against the state-of-the-art method SplaTAM while achieving better results in rendering novel views. As shown in Fig.\ref{fig6}, our rendered images of novel views have fewer artifacts and occlusions. This is because we directly shade the triangular faces in the mesh without optimizing the blank spaces. Moreover, we evaluate the geometric reconstruction of the scene by comparing the rendered depth against the ground-truth. Our method achieves better accuracy in both novel view and training view synthesis. Fig.\ref{fig10} also shows the visual results of our reconstruction on ScanNet++ dataset.


\subsection{Discussion}
\subsubsection{Mesh Reconstruction from LiDAR Scans}
In addition to depth map, our presented method can directly deal with LiDAR point clouds. We conduct experiments on the Newer College Dataset, where the point clouds and their corresponding images are treated as input. As shown in Fig.\ref{fig9}, we compared the reconstruction results with Puma. Our method achieved similar reconstruction accuracy to Puma but with more complete results. The promising results indicate that our presented method is able to handle the LiDAR scan as well. Note that our result may contain invalid triangles in blank regions due to the noisy observations.

\begin{figure}[t]
	\centering
	\includegraphics[width=0.48\textwidth]{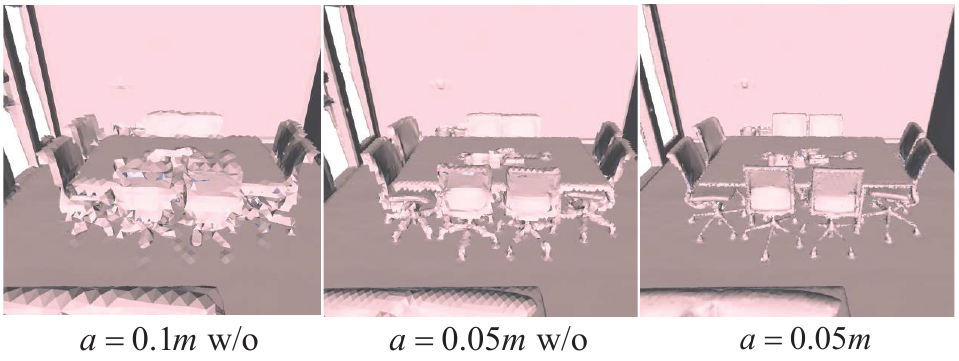}
	\caption{ \textbf{Reconstruction results with different voxel sizes $\alpha$.} $w/o$ means the results without the refinement process. }
	\label{fig11}
\end{figure}

\begin{table}[t]
\centering
\scriptsize
\setlength{\tabcolsep}{12.5pt}

\begin{tabular}{lccc}
\toprule
\textbf{Methods} & \textbf{Params}↓ & \textbf{Iteration FPS}↑ & \textbf{Runtime}↓ \\
\midrule
NICE-SLAM                   & 58.95 K        & 0.57        & 61.5 min \\
Vox-Fusion                  & \cellcolor{tabthird}54.28 K       & \cellcolor{tabthird}0.79        & \cellcolor{tabthird}42.5 min \\
GO-Surf                     & \cellcolor{tabsecond}3.11 K        & \cellcolor{tabsecond}11.11       & \cellcolor{tabsecond}7.0 min    \\
Ours                        & \cellcolor{tabfirst}\textbf{0}        & \cellcolor{tabfirst}\textbf{16.67}       & \cellcolor{tabfirst}\textbf{5.0 min}    \\
\bottomrule
\end{tabular}
\caption{ \textbf{Performance comparison on Replica/Office 0.}  }
\label{tab3}
\end{table}

\subsubsection{Efficiency}
Differently from implicit surface reconstruction, our approach does not require time-consuming sampling, which leads to faster runtime. Tab.\ref{tab3} compares the model parameters, iteration FPS, and runtime. The runtime refers to the total time of running the model, including data loading and processing, while the iteration FPS refers to the processing time of a single frame on the GPU. Our method do not require any model parameters and outperforms the baseline methods on all metrics. NICE-SLAM consists of three hierarchical feature grids, which require costly query operations for reconstruction. Vox-Fusion adopts a standard octree structure, which results in longer query time compared to our hybrid voxel-octree. Although GO-Surf has a fast iteration speed, it requires the traditional Marching Cubes algorithm to extract the mesh. This requires more computational power. Our method only adjusts the surface triangles, avoiding expensive and time-consuming sampling. Thus, it is quite efficient in practice. 
\subsubsection{Ablation Study}
We conduct ablation studies to demonstrate the impact of different branches and parameters of the hybrid voxel-octree.

We compare the impact of different voxel sizes and levels on the hybrid voxel-octree. Tab.\ref{tab4} shows the experimental results on Replica dataset. It can be clearly seen that the reconstruction accuracy increases as the voxel size decreases. However, smaller voxels lead to an exponential growth in the number of leaf nodes in the octree, which significantly increases the processing time and memory usage. Moreover, the hierarchical voxels have similar reconstruction accuracy as the traditional octree with deeper levels, which have shorter insertion time. This indicates that our underlying structure can better trade-off between efficiency and accuracy.

Finally, we examine the effectiveness of different refinement branches. As shown in Tab.\ref{tab5}, the point-based refinement plays a crucial role in significantly improving the accuracy. This is because we directly use the target point cloud extracted from the octree to supervise the position of triangle vertices. The shading-based refinement is supervised by images, which may not provide accurate geometric information. The shading-based refinement primarily focuses on learning the shading model and can be jointly optimized with the point refine branch. Fig.\ref{fig11} shows the reconstruction results with various settings.

\begin{figure}[t!]
	\centering
	\includegraphics[width=0.48\textwidth]{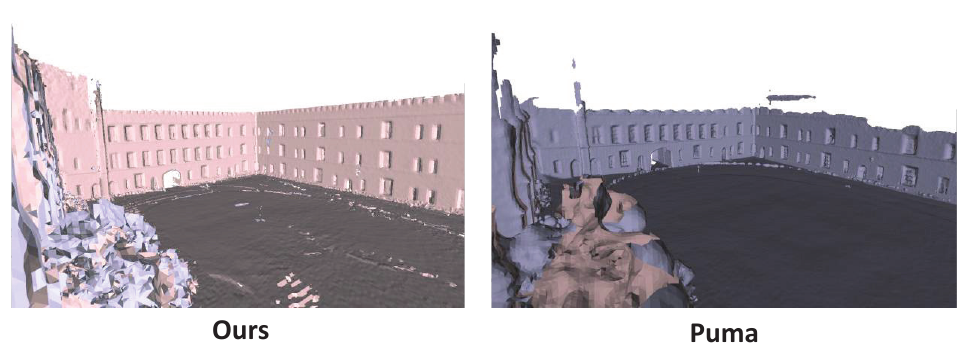}
	\caption{ \textbf{Reconstruction results on the Newer College Dataset.} }
	\label{fig9}
\end{figure}

\begin{table}[t!]
\centering
\scriptsize
\setlength{\tabcolsep}{7.5pt}

\begin{tabular}{lccc}
\toprule
\textbf{Voxel (size {[}m{]} / level)} & \textbf{Acc {[}cm{]}} & \textbf{Insertion time} & \textbf{Memory usage} \\
\midrule
0.1 / 3                & 0.92       & 0.18 s       & 13.35 MiB                \\
0.05 / 3               & 0.59       & 0.32 s       & 21.40 MiB              \\
0.02 / 3               & 0.44       & 1.03 s       & 59.13 MiB                \\   
0.02 / 1               & 0.58       & 1.03 s       & 19.07 MiB               \\  
\bottomrule
\end{tabular}
\caption{ \textbf{Performance analysis on different voxel sizes and levels.} Insertion time means the time required to insert a single poind cloud (819,200 points) into the leaf nodes. Memory usage means the amount of CPU memory utilized to store a frame of point cloud. }
\label{tab4}
\end{table}

\begin{table}[t!]
\centering
\scriptsize
\setlength{\tabcolsep}{9pt}

\begin{tabular}{ccccc}
\toprule
\textbf{Point-based} & \textbf{Shading-based} & \multirow{2}{*}{\textbf{ Acc {[}cm{]}↓}} & \multirow{2}{*}{\textbf{NC↑}} & \multirow{2}{*}{\textbf{F-score↑}} \\
\textbf{Refinement} & \textbf{Refinement} \\
\midrule
$\times$                       & $\times$                         & 0.96          & 0.93 & 0.87     \\
$\checkmark$                      & $\times$                        & 0.61          & 0.95 & 0.92     \\
$\times$                      & $\checkmark$                        & 0.95          & 0.93 & 0.88     \\
$\checkmark$                      & $\checkmark$                        & \textbf{0.59}          & \textbf{0.96} & \textbf{0.93}     \\
\bottomrule
\end{tabular}
\caption{ \textbf{Refinement branch Ablation on Replica/Office 3}. }
\label{tab5}
\end{table}

\section{Conclusion}
We introduce HVOFusion, incremental mesh reconstruction using hybrid voxel-octree. The hybrid voxel-octree effectively fuse octree with voxel structures so that we can take advantage of both implicit surface and explicit triangular mesh representation. This structure allows the partial reconstruction of the scene geometry during the octree construction. To enhance the reconstruction accuracy, we designed point-based and shading-based refinement branch to recover surface geometry and vertex colors. Experiments show that our method can quickly and accurately reconstruct a scene with realistic colors. One limitation of our method is that it is sensitive to the quality of the input point cloud and has difficulty reconstructing unobserved regions. We intend to enhance the completeness and robustness of the model in future work.

\section*{Acknowledgments}
This work is supported by National Natural Science Foundation of China under Grants (62376244). It is also supported by Information Technology Center and State Key Lab of CAD\&CG, Zhejiang University.

\bibliographystyle{named}
\bibliography{ijcai24}

\end{document}